\documentclass[12pt]{article}
\usepackage{framed,epsf,latexsym,amsmath,
amssymb,amscd,mathrsfs}
\usepackage{enumitem}                                 
\usepackage{graphicx}
\usepackage{algorithm}
\usepackage{algorithmic}
\usepackage[pdftex,colorlinks]{hyperref}
\usepackage[margin=1.0in]{geometry}
\usepackage{color}
\usepackage{amsmath,amssymb,graphicx}

\usepackage[round,comma]{natbib}
\newtheorem{theorem}{Theorem}[section]

\newtheorem{counter-example}[theorem]{Counter example}
\newtheorem{open question}[theorem]{Open question}

\newtheorem{claim}{Claim}

\newcommand{\proof}{{\par\noindent {\bf Proof}\space\space}}

\newcommand{\ignore}[1]{}

\newcommand{\ct}{{\cal T}}

\newcommand{\ch}{{\cal H}}

\newcommand{\cw}{{\cal W}}

\newcommand{\cy}{{\cal Y}}
\newcommand{\cz}{{\cal Z}}

\newcommand{\bz}{{\bold{z}}}

\DeclareMathOperator*{\argmax}{argmax}

\newcommand{\inner}[1]{\langle #1 \rangle}

\newcommand{\reals}{{\mathbb R}}

\newcommand{\complex}{{\mathbb C}}

\newcommand{\proofbox}{\begin{flushright}$\Box$\end{flushright}}

\DeclareMathOperator{\Err}{Err}
\DeclareMathOperator{\BErr}{B-Err}

\DeclareMathOperator{\Ldim}{L}
\DeclareMathOperator{\BLdim}{BL}
\DeclareMathOperator{\POB}{POB}

\title{The price of bandit information in multiclass online classification}

\author{Amit Daniely\thanks{Department of Mathematics, Hebrew
                 University, Jerusalem 91904, Israel.  amit.daniely@mail.huji.ac.il}
 \and Tom Halbertal\thanks{Department of Mathematics, Hebrew
                 University, Jerusalem 91904, Israel.  tom.halbertal@mail.huji.ac.il}}

\begin{document}

\maketitle
\begin{abstract}
We consider two scenarios of multiclass online learning of a hypothesis class $H\subseteq Y^X$. In the {\em full information} scenario, the learner is exposed to instances together with their labels. In the {\em bandit} scenario, the true label is not exposed, but rather an indication whether the learner's prediction is correct or not. We show that the ratio between the error rates in the two scenarios is at most $8\cdot|Y|\cdot \log(|Y|)$ in the realizable case, and $\tilde{O}(\sqrt{|Y|})$ in the agnostic case. The results are tight up to a logarithmic factor and essentially answer an open question from \cite{DanielySaBeSh11}.

We apply these results to the class of multiclass linear classifiers in $\reals^d$ with margin $\frac{1}{D}$. We show that the bandit error rate of this class is $\tilde{\Theta}\left(D^2|Y|\right)$ in the realizable case and $\tilde{\Theta}\left(D\sqrt{|Y|T}\right)$ in the agnostic case. This resolves an open question from \cite{KakadeShTe08a}.

\end{abstract}

\section{Introduction}
Online multiclass classification is an important task in Machine Learning.
In its basic form, which we refer as the {\em full information} scenario, the learner is required to predict the label of a new example, based on previously observed labeled examples.
Recently, the {\em bandit} scenario has received much attention (e.g. \cite{AuerCeFrSc03,KakadeShTe08a,DaniHaKa08,AuerCeFi02}). Here, the learner does not observe labeled examples, but rather, it observes unlabeled examples, predicts their labels and only receives an indication whether his prediction was correct. The relevance of the bandit scenario to practice is evident -- a canonical example is internet advertising, where the advertiser chooses a commercial (which is thought as a label) upon the information it has on the user (which is thought as an instance). After choosing a commercial, the advertiser only knows if the user has clicked the commercial or not.

Let $X$ be an instance space and $Y$ a label space. Denote $k=|Y|$. To evaluate learning algorithms, it is common to compare them to the best hypothesis coming from some fixed hypothesis class $H\subseteq Y^X$. 
We define the {\em error rate} of $H$ as the least number, $\Err_H(T)$, for which {\em some} algorithm is guaranteed to make at most $\Err_H(T)$ mistakes more than the best hypothesis in $H$, when running on a sequence of length $T$.
We emphasize that we consider {\em all} algorithms, not only efficient ones.

It is clear that learning is harder in the bandits scenario.
The purpose of this work is to quantify how larger is the error rate in this scenario. Our main results show that, for every hypothesis class $H$, the error rate in the bandit scenario is only $\tilde{O}(k)$ times larger in the realizable case (i.e. in the case that some hypothesis in $H$ makes no mistakes) and $\tilde{O}(\sqrt{k})$ times larger in the general (agnostic) case. We note that our results hold also for the multiclass multi-label categorization, where a set of labels are allowed to be correct.
As an application, we use our results to quantify the error rate of the class of large margin halfspaces classifiers.

\subsection{Related Work}
{\bf Cardinality based vs. Dimension based bounds.} The celebrated result of \cite{LittlestoneWa89b} shows that, in the full-info scenario, the error rate is upper bounded by $O(\sqrt{\log(|H|)T})$. In the full-info-realizable case, the majority algorithm achieves an error rate of $O(\log(|H|))$.

These two bounds are tight for several hypothesis classes. However, there are several important classes for which much better error rates can be achieved. For example, those bounds are meaningless for infinite hypothesis classes. However, several such classes (e.g. the class of large margin halfspaces classifiers) do admit a finite error rate. 

The reason to those deficiencies is that the quantity $\log(|H|)$ does not quantify the true complexity of the class, but only upper bounds it.
To remedy that, \cite{DanielySaBeSh11}, following
a binary version from \cite{Ben-DavidPaSh09} and \cite{Littlestone88}, proposed a notion of
dimension (a-la VC dimension), called the {\em Littlestone
  dimension}. As shown in \cite{DanielySaBeSh11}, the error rate of $H$, in the full-info scenario, is $\tilde{\Theta}(\sqrt{\Ldim(H)T})$ in the agnostic case and $\Theta(\Ldim(H))$ in the realizable case\footnote{A detailed study of online analogs to statistical complexity measures can be found in \cite{RakhlinSrTe10}).}.

The results of \cite{DanielySaBeSh11} show that the term $\log(|H|)$ in the result of \cite{LittlestoneWa89b} and in the bound of the majority algorithm can be replaced by $\Ldim(H)$ (the algorithms they use are different, however), leading to a tight (up to $\log$ factors of $T$ and $k$) characterization of the error rate.
Our results can be seen as analogues of these results in the bandit scenario:
By the algorithm of \cite{AuerCeFrSc03}, the error rate of $H$ in the bandit scenario is $O(\sqrt{kT\log(|H|)})$. By the Majority algorithm, the error rate in the realizable-bandit scenario is $O(k\log(|H|))$. Our results upper bound the error rates by $\tilde{O}(\sqrt{kT\Ldim(H)})$ and $\tilde{O}(k\Ldim(H))$ respectively.

Since the Littlestone dimension characterizes the full-info error rate, our results imply an upper bound on the ratio between the bandit and full-info error rates. To the best of our knowledge, these are the first upper bounds on this ratio that hold for every class.  
We note also that since $\Ldim(H)\le \log(|H|)$, our bounds, up to $\log$ factors, imply the bounds of \cite{AuerCeFrSc03} and the majority algorithm.

{\bf Comparison to other settings.}  In the statistical/PAC settings, one
assumes that the sequence of examples is drawn i.i.d. from some
distribution on $X\times Y$. In these settings, it is not hard to show
that the bandit error rate is at most $O(k)$ times larger than the
full-info rate (see Section \ref{sec:conclusion} and \cite{DanielySaBeSh11}). 
Our results generalize these facts to the adversarial setting.

\section{Our Results}
\subsection{Problem setting and Background}
{\bf Setting.} Fix an instance space $X$ and a label space
$Y$. Denote $Z=X\times Y$, $\cy =2^Y$, $\cz=X\times \cy$ and
$k=|Y|$. We consider two scenarios of multiclass online learning. In
the {\em full information} scenario, at each step $t=1,2,\ldots$ a
{\em full-info learning algorithm} is exposed to an instance $x_t\in
X$, predicts a label $\hat y_t\in Y$ and then observes a list of true
labels $Y_t\subset Y$ (note that this is little more general than the vanilla multiclass setting in which $|Y_t|=1$). The prediction $\hat y_t$ can be based only on the previously observed labeled examples $(x_1,Y_1),\ldots,(x_{t-1},Y_{t-1})$ and on $x_t$.
The {\em bandit} scenario is similar. The sole difference is that,
after a {\em bandit learning algorithm} predicts a label, the true
labels are not exposed, but only an indication whether the algorithm's
prediction was correct or not. Therefore, the prediction $\hat y_t$
can be based only on the previously observed unlabeled examples
$x_1,\ldots,x_{t-1},x_t$ and on previously obtained indications
\mbox{$1(\hat y_1\in Y_1),\ldots,1(\hat y_{t-1}\in Y_{t-1})$}. We assume that
the choice of the sequence $(x_t,Y_t)$ is adversarial, but the
adversary chooses $Y_t$ before the algorithm predicts $\hat y_t$. In
particular, the algorithm may choose $\hat y_t$ at random after the
adversary chose $Y_t$.

Let $H$ be a hypothesis class, which might be either class of functions from $X$ to $Y$, or, a class of functions from $X$ to $\mathbb R^Y$.  We say that a sequence $(x_1,y_1)\ldots,(x_T,y_T)\in Z$ is {\em realizable} by $H$ if there exists a function $h\in H$ such that either $\forall 1\le t\le T,\;h(x_t)= y_t$, for the case that $H\subset Y^X$, or $\forall 1\le t\le T,\;h_{y_t}(x_t)\ge 1+\max_{y\ne y_t}h_{y}(x_t)$, in the case\footnote{In some contexts 
it favourable to use a margin-dependent notion of realization. Namely, to define a {\em $\gamma$-realizable sequence} by requiring that $\forall 1\le t\le T$, $h_{y_t}(x_t)\ge \gamma+\max_{y\ne y_t}h_{y}
(x_t)$. Observing that a sequence is $\gamma$-realizable by $H$ iff it
is realizable by $\left(\frac{1}{\gamma}\cdot H\right)$, it is easy to
interpolate between the two definitions. Our choice of the above
definition is merely for the sake of clarity.} that $H\subset (\reals^Y)^X$. We denote by $H(T)\subset Z^T$ the sequences of length $T$ that are realizable by $H$.
We say that a sequence $(x_1,Y_1),\ldots,(x_T,Y_T)\in \cz$ is {\em realizable by H} if there exist $y_1\in Y_1,\ldots,y_T\in Y_T$ such that the sequence $(x_1,y_1),\ldots,(x_T,y_T)$ is realizable by $H$.

The {\em error of $H$ on a sequence $\bz=((x_1, Y_1),\ldots,(x_T,Y_T))\in \cz^T$} is the minimal number of errors that a hypothesis from $H$ makes on the sequence $\bz$. Namely,
$$\Err(H,\bz)=\min_{((x_1, y_1),\ldots,(x_T,y_t))\in H(T)}\sum_{t=1}^T1(y_t\not\in Y_t) ~.$$

Let $A$ be a (either full-info or bandit) learning algorithm. Given $\bz\in \cz^T$, we denote by $\Err(A,\bz)$ the expected number of errors $A$ makes, running on the sequence $\bz$. We define the {\em realizable error rate} of $A$ w.r.t. $H$ as the worst case performance of $A$ on a length $T$ realizable sequence, namely,
$$\Err_{A}^r(T)=\sup_{\bz\in \cz^T,\;\Err(H,\bz)=0}\Err(A,\bz)~.$$
The {\em agnostic error rate} of $A$ is its worst case performance over all length $T$ sequences, namely,
$$\Err_{A}^a(T)=\sup_{\bz\in \cz^T}\Err(A,\bz)-\Err(H,\bz)~.$$
The {\em realizable and agnostic full-info error rates} of the class $H$ are the best achievable error rates, namely,
$$\Err^r_H(T)=\inf_{A\text{ is a full-info alg.}}\Err_A^r(T)
\text{ and }
\Err^a_H(T)=\inf_{A\text{ is a full-info alg.}}\Err_A^a(T)~.$$
Similarly, the {\em realizable and agnostic bandit error rates} of the class $H$ are 
$$\BErr^r_H(T)=\inf_{A\text{ is a bandit alg.}}\Err_A^r(T)
\text{ and }
\BErr^a_H(T)=\inf_{A\text{ is a bandit alg.}}\Err_A^a(T)~.$$
Our main focus is to understand how larger is the error rate in the bandit scenario, compared to the full-info scenario. Thus, we define the agnostic and realizable {\em price of bandit information} of $H$ by
$$\POB^r_H(T)=\frac{\BErr^r_H(T)}{\Err^r_H(T)}\text{ and }\POB^a_H(T)=\frac{\BErr^a_H(T)}{\Err^a_H(T)}~.$$

{\bf The Littlestone dimensions.} \cite{DanielySaBeSh11} (following
\cite{Ben-DavidPaSh09} and \cite{Littlestone88}) defined two
combinatorial notions of dimension that characterize the error rates of
a class $H$. Let $\ct$ be a rooted tree whose internal nodes are labeled
by $X$ and whose edges are labeled by $Y$. We say that $\ct$ is {\em
  L-shattered} by $H$ if, for every root-to-leaf path
$x_1,\ldots,x_T$, the sequence $(x_1,y_1),\ldots,(x_{T-1},y_{T-1})$,
where $y_t$ is the label associated with the edge $x_t \to x_{t+1}$, is realizable by $H$. The {\em Littlestone dimension} of $H$, denoted $\Ldim(H)$, is the maximal depth of a complete binary tree\footnote{By a complete binary tree, we mean a tree whose all internal nodes have two children and all leaves are at the same depth.} that is L-shattered by $H$. 
We say that $\ct$ is {\em BL-shattered} by $H$ if, for every root-to-leaf path $x_1,\ldots,x_T$, if $y_t$ is the label associated with $x_t\to x_{t+1}$, then there exists a realizable sequence $(x_1,y'_1),\ldots,(x_{T-1},y'_{T-1})$ such that $\forall i,\;y'_t\ne y_t$. The {\em Bandit Littlestone dimension} of $H$, denoted $\BLdim(H)$, is the maximal depth of a complete $k$-ary tree that is BL-shattered by $H$.

\begin{theorem}\label{thm:Littlestone_dimensions}(\cite{DanielySaBeSh11})
\begin{itemize}
\item For every class $H$ and for every $T\ge \Ldim(H)$, $\frac{1}{2}\Ldim(H)\le \Err^r_H(T)\le \Ldim(H)$ and 
$\Omega\left(\sqrt{\Ldim(H)T}\right)\le \Err^a_H(T)\le O\left(\sqrt{\Ldim(H)T\log(kT)}\right)$.
\item For every class $H$, $\BErr^r_H(T)\le \BLdim(H)$. Moreover, for every deterministic bandit algorithm, $\Err^r_A(T)\ge \min\{T,\BLdim(H)\}$.
\end{itemize}
\end{theorem}

{\bf The class of large-margin multiclass linear separators.} Denote by $B^d$ the unit ball in $\mathbb R^d$. We identify every matrix $W\in M_{k\times d}(\mathbb R)$ with the linear function it defines on $B^d$ (i.e. $x\mapsto Wx$). Denote by $\|W\|_F$ the Frobenius norm of $W$, namely, \mbox{$\|W\|_F=\sqrt{\sum_{i=1}^k\sum_{j=1}^dW_{ij}^2}$}.
For $D>0$ let $\cw^{d,k}(D)=\{W\in M_{k\times d}(\mathbb R):\|W\|_F\le D\}$.  

A multiclass variant of the Perceptron algorithm (e.g. section 5.12 in \cite{DudaHa01}) makes at most $2\cdot D^2$ mistakes whenever it runs on a sequence that is realizable by $\cw^{d,k}(D)$. Therefore, $\Ldim(\cw^{d,k}(D))\le 2\cdot D^2$. Also, it is not hard to see that $\min\{d,\lfloor D^2\rfloor\}\le \Ldim(\cw^{d,k}(D))$. Thus, we have
\begin{equation}\label{eq:Ldim_linear}
\min\{d,\lfloor D^2\rfloor\}\le \Ldim(\cw^{d,k}(D))\le 2\cdot D^2
\end{equation}

\subsection{Results}
Our first result bounds the bandit-realizable error rate in terms of the Littlestone dimension.
\begin{theorem}\label{thm:main_realizable}
For every hypothesis class $H$,  $\BErr_H^r(T)\le 4k\log(k)\Ldim(H)$. Moreover, the upper bound is achieved by a deterministic algorithm.
\end{theorem}
Together with Theorem \ref{thm:Littlestone_dimensions} we conclude that the bandit-realizable error rate is at most $\tilde{O}(k)$ larger than the full-info-realizable error rate. Namely, for every hypothesis class $H$,
\begin{equation}\label{eq:POB_realizable}
\POB^r_H(T)\le  8 k\cdot \log (k)~.
\end{equation}
It is not hard to see (e.g. by Claim \ref{lem:prob_lem}) that for finite $X$, $H=Y^X$ and $T\ge (k-1)\cdot |X|$, we have that $\BErr_{H}^r(T)\ge\frac{(k-1)\cdot |X|}{2}=\frac{(k-1)\cdot \Ldim(H)}{2}$. Thus, Theorem \ref{thm:main_realizable} is tight  up to a factor of $\log(k)$. By Theorem \ref{thm:Littlestone_dimensions},
$$\POB^r_H(T)=\frac{\BErr_H^r(T)}{\Err_H^r(T)}\ge \frac{\frac{(k-1)\cdot \Ldim(H)}{2}}{\Ldim(H)}=\frac{(k-1)}{2}~.$$ 
Thus, Equation (\ref{eq:POB_realizable}) is tight up to a factor of $\log(k)$ as well.
In \cite{DanielySaBeSh11} it was asked how large the ratio $\frac{\BLdim(H)}{\Ldim(H)}$ can be. It can be easily seen that $\frac{\BLdim(H)}{\Ldim(H)}$ can be as large as $k-1$ (this is true, for example, when $X$ is finite and $H=Y^X$). Theorem \ref{thm:main_realizable}, together with Theorem \ref{thm:Littlestone_dimensions}, shows that $\frac{\BLdim(H)}{\Ldim(H)}\le 4k\log(k)$, essentially answering the question of \cite{DanielySaBeSh11}.

For the agnostic case we show the following result:
\begin{theorem}\label{thm:main_agnostic}
For every class $H$,  $\BErr_H^a(T)\le e\cdot\sqrt{Tk\Ldim(H)\log(T\cdot k)}$.
\end{theorem}
Together with Theorem \ref{thm:Littlestone_dimensions}, it follows that for every class $H$,
\begin{equation}\label{eq:POB_agnostic}
\POB^a_H(T)=O\left(\sqrt{k\cdot \log(k\cdot T)}\right)~.
\end{equation}
Relying on the construction from section 5 of \cite{AuerCeFrSc03}, it is not hard to show that for $H=Y^X$ and $T\ge k\cdot |X|=k\cdot \Ldim(H)$ it holds that $\BErr^a_H(T)\ge \frac{1}{20}\sqrt{\Ldim(H)\cdot T\cdot k}$.
Thus, together with Theorem \ref{thm:Littlestone_dimensions}, Theorem \ref{thm:main_agnostic} and Equation (\ref{eq:POB_agnostic}) are tight up to a logarithmic factor of $\log(k\cdot T)$.

Next, we apply Theorems \ref{thm:main_realizable} and \ref{thm:main_agnostic} to analyse the bandit error rate of large margin multiclass linear separators.
\begin{theorem}\label{thm:BLdim_linear_margin}
For every $D>0$ and $d,k\in\mathbb N$
$$\BErr_{\cw^{d,k}(D)}^r(T)\le 8\cdot k\cdot \log(k)\cdot D^2,\;\;
\BErr_{\cw^{d,k}(D)}^a(T)\le 4\cdot D\cdot\sqrt{Tk\log(T\cdot k)}$$
Moreover, for $L=\min\{d,\lfloor D^2 \rfloor\}$ and $T\ge k\cdot L$,
$$\BErr_{\cw^{d,k}(D)}^r(T)\ge \frac{(k-1)\cdot L}{2}, \;\; 
\BErr_{\cw^{d,k}(D)}^a(T)\ge \frac{1}{20}\sqrt{LTk}$$
\end{theorem}

\cite{KakadeShTe08a} have shown an (inefficient) randomized algorithm that makes, w.p. $1-\delta$, at most $O\left( k^2 D^2\ln\left(\frac{T+k}{\delta}\right)\cdot\left(\ln D+\ln\ln \left(\frac{T+k}{\delta}\right)\right)\right)$ mistakes, whenever it runs on a sequence $(x_1,Y_1),\ldots,(x_T,Y_T)$ that is realizable by $\cw^{d,k}(D)$. It has been asked there what is the optimal error rate, and whether there exists an asymptotically finite bound on the error rate that does not depend on the dimension $d$. Theorem \ref{thm:BLdim_linear_margin} answers the second question in the affirmative and essentially answers the first question. Also, \cite{KakadeShTe08a} have conjectured that for fixed $D$ and $k$, the bandit agnostic error rate of $\cw^{d,k}(D)$ should be $O(\sqrt{T})$. Theorem \ref{thm:BLdim_linear_margin} validates this conjecture, up to a factor of $\sqrt{\log(T)}$.

The bound in Theorem \ref{thm:BLdim_linear_margin} is rather tight when the dimension, $d$, is larger than the complexity $D^2$. To complete the picture, we note that in \cite{KakadeShTe08a} it has been shown that $\BErr_{\cw^{d,k}(D)}^r(T)\le O(k^2d\log(D))$. Here we show a corresponding lower bound. 
\begin{theorem}\label{thm:linear_no_margin}
For every $D^2\ge {k^3 d}$ and $T\ge  \frac{dk^2}{8}$,
$\BErr_{\cw^{d,k}(D)}^r(T)\ge \lfloor d/2\rfloor \cdot (k-1)\cdot \frac{k}{4}$.
\end{theorem}

\subsection{Proof techniques}
The proof of Theorem \ref{thm:main_realizable} constitutes most of the technical novelty of the paper.
The algorithm we use belongs to the family of ``majority vote"
algorithms such as the Standard Optimal Algorithms of \cite{Littlestone88}, \cite{Ben-DavidPaSh09}
and \cite{DanielySaBeSh11}. These algorithms start with a hypothesis
class $H_1=H$. At each step $t$, they predict the label predicted by
``most" hypotheses in $H_t$, where ``most" is quantified in a certain way. After an indication is given for that prediction (i.e., after the true label is exposed in the full-info scenario or after an indication whether the algorithm's guess was correct or not in the bandit scenario), the algorithm constructs $H_{t+1}$ by throwing away all functions that are in contradiction with that indication.

A crucial distinction is that instead of a single hypothesis class, our algorithm keeps a {\em collection} of hypothesis classes. At each step, each class in that collection is either splited, thrown away, or remains untouched. The prediction at each step aims to minimize a measure of capacity for collections of hypothesis classes, which we define. We show that this measure shrinks to $1$ after $4k\cdot \log(k)\cdot\Ldim(H)$ mistakes. From that point, the algorithm makes no further mistakes.

Theorem \ref{thm:main_agnostic} is based on an argument from \cite{Ben-DavidPaSh09} (see also \cite{DanielySaBeSh11}). We represent each class by a relatively small number of experts and apply the result of \cite{AuerCeFrSc03} on this set of experts. Theorem \ref{thm:BLdim_linear_margin} is deduced from Theorems \ref{thm:main_realizable}, \ref{thm:main_agnostic} and Equation (\ref{eq:Ldim_linear}).

To prove Theorem \ref{thm:linear_no_margin}, we first consider the class $H$ of all functions \mbox{$f:\left[\lfloor\frac{d}{2}\rfloor\right]\times [k]\to [k]$} such that $f|_{\{j\}\times [k]}$ 
is a bijection for every $j\in \left[\lfloor\frac{d}{2}\rfloor\right]$. We show that \mbox{$\BErr_{H}^r(T)\ge \lfloor\frac{d}{2}\rfloor \cdot (k-1)\cdot \frac{k}{4}$}. Then, we adapt a construction from \cite{DanielySaBeSh11} to show that $H$ can be realized by $\cw^{d,k}(D)$.

\section{Proofs}
Throughout, we denote by $k$ the number of labels.
We prove Theorems \ref{thm:main_realizable} and \ref{thm:main_agnostic} for the case that $H\subset Y^X$. The case of real-valued $H$ can be handled along the same lines. We say that a hypothesis class $H_1$ is {\em realized} by $H_2$, if, $\forall T,H_1(T)\subset H_2(T)$. It is clear that in this case the error rates and the Littlestone dimensions of $H_1$ are no larger than those of $H_2$.

\subsection{Theorem \ref{thm:main_realizable}}
Let $\mathcal{H}$ be a collection of non-empty subsets of $H$.
We define its {\em capacity} by \mbox{$C(\mathcal{H})=\sum_{V \in
  \mathcal{H} }k^{2L(V)}$}. We note that for $\mathcal{H}=\{H\}$,
it holds that $C(\mathcal{H})=k^{2L(H)}$. Also, for non-empty $\ch$, $C(\mathcal{H})\ge 1$.
Our algorithm starts with $\ch_1=\{H\}$. At each step it modifies $\ch_t$ such that (1) $C(\ch_t)$ shrinks with every mistaken prediction and (2) all hypotheses that are consistent with the previously observed instances are in one of the subclasses of $\ch_t$ .

Given $x \in X$, $y \in Y$ and $V \subset H$ devote $V_x^y=\{f \in V:f(x)=y\}$. For a collection, $\ch$, of subsets of $H$ we define
$$\Lambda(\mathcal{H},x,y_0 )=\{V \in \mathcal{H}:\forall y \neq y_0, \Ldim (V_x^y )<L(V)\}$$
$$\lambda(\mathcal{H},x,y_0 )=\{V_x^y:V \in \Lambda(\mathcal{H},x,y_0 ),\; y\ne y_0,\; V_x^y\ne\emptyset\} \cup \mathcal{H} \setminus \Lambda(\mathcal{H},x,y_0 )$$
$$P_{\mathcal{H},x} (y)=C(\mathcal{H})-C(\lambda(\mathcal{H},x,y))$$

\begin{algorithm}[th] 
\caption{} \label{algo:main}
\begin{algorithmic}[1]
\STATE Set $\mathcal{H}_1= \{H\}$.
\FOR {$t=1,2, \ldots$}
\STATE receive $x_t$
\STATE Predict $y \in argmax_{y \in Y} P_{\mathcal{H}_t,x_t } (y)$.
\STATE If the prediction is wrong, update $\mathcal{H}_{t+1}= \lambda (\mathcal{H}_t,x_t,y)$. Otherwise, $\ch_{t+1}=\ch_t$.
\ENDFOR
\end{algorithmic}
\end{algorithm}

\begin{claim}
Algorithm \ref{algo:main} makes less than $4\Ldim(H) k\log(k)$ mistakes.
\end{claim}
\proof
Fix $V \in \mathcal{H}_t$ and let $y \in argmax_{y' \in Y} \Ldim(V_x^{y'})$. Since there is at most one $y'\in Y$ such that $\Ldim(V_x^{y'})=\Ldim(V)$ (see \cite{Littlestone88}), it follows that for every $y' \neq y$, $\Ldim(V_x^{y'})<\Ldim(V)$. In particular, $V\in \Lambda(\ch_t,x,y)$ and
\begin{eqnarray*}
k^{2\Ldim(V)}-\sum_{y'\mid V_x^{y'}\ne\emptyset ,\;y'\ne y}k^{2\Ldim(V^{y'}_x)} &\ge& k^{2\Ldim(V)}-(k-1)\cdot k^{2\Ldim(V)-2}
\\
&=& \left(1-\frac{k-1}{k^2}\right)k^{2\Ldim(V)}
\\
&\ge & \left(1-\frac{1}{k}\right)k^{2\Ldim(V)}~.
\end{eqnarray*}
Thus,
\begin{eqnarray*}
\sum_{y\in Y} P_{\mathcal{H}_t,x_t}(y) &=& \sum_{y \in Y} \sum_{V \in \Lambda(\mathcal{H}_t,x_t,y)}(k^{2L(V)} -\sum_{y'\mid V_x^{y'}\ne\emptyset ,\;y'\ne y}k^{2L(V_{x_t}^{y'})})
\\
&\ge & \sum_{V \in \mathcal{H}_t}\sum_{y \in Y:V \in \Lambda(\mathcal{H}_t,x_t,y)}(k^{2L(V)}- \sum_{y'\mid V_x^{y'}\ne\emptyset ,\;y'\ne y}k^{2L(V_{x_t}^{y'})})
\\
&\ge & \sum_{V \in \mathcal{H}_t}\left(1-\frac{1}{k}\right)k^{2\Ldim(V)}
\\
&=& \left(1-\frac{1}{k}\right)C(\ch_t)
\end{eqnarray*}
It follows that for some $y\in Y$, $P_{\mathcal{H}_t,x_t }(y) \ge \frac{1}{k}\left(1-\frac{1}{k}\right)C(\ch_t)\ge\frac{1}{2k}C(\ch_t)$. Thus, if the algorithm errs at time $t$ then
$$ C(\mathcal{H}_t) - C(\mathcal{H}_{t+1}) \ge \frac{1}{2k}C(\mathcal{H}_t )\Rightarrow C(\mathcal{H}_{t+1})\le \left(1-\frac{1}{2k}\right)C(\ch_t)$$
It follows that after $4\Ldim(H) k\log(k)$ mistakes it will hold that
\begin{eqnarray*}
C(\ch_t) &\le & \left(1-\frac{1}{2k}\right)^{4\Ldim(H)k\log(k)}C(\ch_1)
\\
&< & e^{-2\Ldim(H)\log(k)}k^{2\Ldim(H)}=1
\end{eqnarray*}
However, it is not hard to see that each hypothesis which is consistent with the history up to time $t-1$ is in one of the classes of $\ch_t$. As we assume that the sequence is realizable, there is at least one consistent hypothesis. Thus, for every $t$, $C(\ch_t)\ge 1$. It follows that the algorithm makes less than $4\Ldim(H) k\log(k)$ mistakes.
\proofbox

\subsection{Theorem \ref{thm:main_agnostic}}
We use a result from \cite{AuerCeFrSc03}, which we briefly
describe next. Suppose that at each step, $t$, before the algorithm chooses its prediction, it observes $N$ {\em advices} $(f_1^t,\ldots,f_N^t)\in
Y^N$, which can be used to determine its prediction. We refer to
$f_i^t$ as the prediction made by the {\em expert} $i$ at time $t$ and
denote by $L_{i,T}=|\{t\in [T]:f_{i,t}\not\in Y_{t}\}|$ the {\em loss}
of the expert $i$ at time $T$. For every sequence $z\in \cz^T$, the algorithm from \cite{AuerCeFrSc03}, section 7, makes at most $\min_{i\in [N]}L_{i,T}+e\sqrt{kT\log(N)}$ mistakes in expectation.

Suppose that, for every $f\in H$, we construct an expert, $E_f$, whose advice at time $t$ is $f(x_t)$. Denote by $L_{f,t}$ the loss of the expert $E_f$ at time $t$. Running the algorithm of \cite{AuerCeFrSc03} with this set of experts yields an algorithm whose agnostic error rate is at most $e\sqrt{kT\log(|H|)}$. We proceed by imitating this set of experts with a more compact set of experts, which will allow us to bound the loss in terms of $\Ldim(H)$ instead of $\log(|H|)$.

Let $A_T = \{A\subset [T]\mid |A|\le\Ldim(H)\}$. For every $A\in A_T$ and $\phi:A\to Y$, we define an expert $E_{A,\phi}$. The expert $E_{A,\phi}$ imitates the SOA algorithm (Algorithm \ref{algo:SOA} in the appendix) when it errs exactly on the examples $\{ x_t \mid t\in A\}$ and the true labels of these examples are determined by $\phi$. The expert $E_{A,\phi}$ proceeds as follows:
\begin{tabbing}
Set $V_1=H$.\\
For \=$t=1,2\ldots,T$\\
\>Receive $x_t$.\\
\>Set $l_t=\argmax_{y \in Y}\Ldim(\{f\in V_t:f(x_t)=y\})$. \\
\>If $t\in A$, Predict $\phi(t)$ and update  $V_{t+1}=\{f\in V_t:f(x_t)=\phi(t)\}$.\\
\> If $t\not\in A$, Predict $l_t$ and update  $V_{t+1}=\{f\in V_t:f(x_t)=l_t\}$.\\
\end{tabbing}
The number of experts we constructed is 
$ \sum_{j=0}^{\Ldim(H)}\binom{T}{j}k^{j}\le (Tk)^{\Ldim(H)}$. Denote the number of mistakes made by the expert $E_{A,\phi}$ after $T$ rounds by $L_{A,\phi,T}$. If we apply the algorithm from \cite{AuerCeFrSc03} with the set of experts we've constructed, we obtain an algorithm that makes at most
$$\min_{A,\phi}L_{A,\phi,T}+e\sqrt{kT\Ldim(H)\log(Tk)}$$
mistakes in expectation, whenever it runs on a sequence $z\in \cz^T$. To finish, we show that $\min_{A,\phi}L_{A,\phi,T}\le \Err(H,z)$.

Let $f\in H$ be a function for which $\Err(H,z)=|\{t\in [T]:f(x_t)\not\in Y_t\}|$. 
Denote by $A\subset [T]$ the set of rounds in which the SOA algorithm errs when running on the sequence $(x_1,f(x_1)),\ldots,(x_T,f(x_T))$ and define $\phi :A\to Y$ by $\phi(t)=f(x_t)$. Since the SOA algorithm makes at most $\Ldim(H)$ mistakes, $|A|\le \Ldim(H)$. It is not hard to see that the predictions of the expert  $E_{A,\phi}$ coincide with the predictions of the expert $E_f$. Thus,
$$L_{A,\phi,T}=|\{t\in [T]:f(x_t)\not\in Y_t\}|=\Err(H,z)~.$$

\subsection{Theorem \ref{thm:BLdim_linear_margin}}
The upper bounds follows from Equation (\ref{eq:Ldim_linear}), together with Theorems \ref{thm:main_realizable} and \ref{thm:main_agnostic}. The lower bounds follows from the corresponding bounds for the class $[k]^{[L]}$, and the fact that this class can be realized by $\cw^{d,k}(D)$: Associate the set $[L]$ with $e_1,\ldots,e_L\in\reals^d$, the first $L$ vectors in the standard basis of $\reals^d$. A function $f:\{e_1,\ldots,e_L\}\to [k]$ can be realized by the matrix $W\in \cw^{d,k}(D)$ whose $i$'th row is $\sum_{j\in [m] : f(j)=i}e_j$.

\subsection{Theorem \ref{thm:linear_no_margin}}
Consider the following game: A r.v., $U$, is sampled uniformly from $Y$. Then the player, that does not observe $U$, try to guess $U$. After each prediction, $\hat y_t$, he only receives an indication whether $U=\hat y_t$.
\begin{claim}\label{lem:prob_lem}
Let $R=|\{1\le t\le |Y|-1:\hat y_t\ne U\}|$. Then $\operatorname{E}[R]\ge \frac{|Y|-1}{2}$.
\end{claim}
\proof
We prove the claim by induction on $|Y|$. For $|Y|=2$ it follows from the fact that $U$ is independent from $\hat y_1$. For $|Y|>2$, we note that, since $U$ is independent from $\hat y_1$, the probability that $\hat y_1\ne U$ is at least $\frac{|Y|-1}{|Y|}$. Also, conditioned on the event that $\hat y_1\ne U$, $U$ and $\hat y_2,\hat y_3,\ldots$ satisfies the requirements of the Claim with the label set $Y\setminus \{\hat y_1\}$. Thus, by the induction hypothesis,
$$\operatorname{E}[R]\ge \left(1+\operatorname{E}[R|\hat y_1\ne U]\right)\cdot\Pr(\hat y_1\ne U) \ge\left(1+\frac{|Y|-2}{2}\right)\cdot \frac{|Y|-1}{|Y|}=\frac{|Y|-1}{2}$$
\proofbox

\begin{claim}\label{claim:perm_class}
Let $H$ be the class of all functions $f:[\Delta]\times [k]\to [k]$ such that $f|_{\{j\}\times [k]}$ is a bijection for every $j\in [\Delta]$. Then
$$\BErr_{H}^r(T)\ge \Delta \cdot (k-1)\cdot \frac{k}{4}$$
\end{claim}
\proof
Consider the following algorithm, applied by the adversary:
\begin{enumerate}[label*=\arabic*.]
\item For $j=1,2\ldots, \Delta$
\begin{enumerate}[label*=\arabic*.]
\item For $m=1,2\ldots, k-1$
\begin{enumerate}[label*=\arabic*.]
\item Choose $y_{j,m}\in [k]\setminus\{y_{j,1},\ldots,y_{j,m-1}\}$ uniformly at random.
\item For $n=1,\ldots k-m$
\begin{enumerate}[label*=\arabic*.]
\item Expose the learner the instance $(j,m)$.
\end{enumerate}
\end{enumerate}
\item Let $y_{j,k}$ be the element in the singleton $[k]\setminus\{y_{j,1},\ldots,y_{j,k-1}\}$.
\end{enumerate}
\end{enumerate}
By Claim \ref{lem:prob_lem}, for every $(j,m)$, the adversary causes the learner to make $\ge\frac{k-m}{2}$ mistakes at the predictions for the instance $(j,m)$. Thus, the expected value of the total number of mistakes is $\Delta\cdot \sum_{m=1}^{k-1}\frac{(k-m)}{2}=\Delta\cdot (k-1)\cdot \frac{k}{4}$. Also, it is clear that the function $f:[\Delta]\times [k]\to [k]$ defined by $f(j,m)=y_{j,m}$ is in $H$. Thus, the sequence produced by the adversary is realizable by $H$.
\proofbox
Theorem \ref{thm:linear_no_margin} follows from the following claim.
\begin{claim}
If $D^2\ge k^3d$ then the class $H$ from Claim \ref{claim:perm_class} with $\Delta=\lfloor \frac{d}{2}\rfloor$ is realized by $\cw^{d,k}(D)$.
\end{claim}
\proof
Let $\bar d=\lfloor d/2\rfloor$. Instead of working in $\reals^d$, we work in $\complex^{\bar d}$. We identify each \mbox{$(j,m)\in [\bar d]\times [k]$} with $x_{j,m}:=e^{\frac{m 2\pi i}{k}}\cdot e_j$. Here, $e_j$ is the $j$'th vector in the standard basis of $\complex^{\bar d}$. 

Let $f\in H$. We must show that $f$ is induced by some $W\in\cw^{d,k}(D)$ in the sense that
$$\forall (j,m)\in [\bar d]\times [k],\;(Wx_{j,m})_{f(j,m)}\ge 1+\max_{m'\ne m} (Wx_{j,m})_{m'}~.$$
Indeed, we let $W\in M_{k,\bar d}(\complex)$ be the matrix defined by
$$\forall (j,m)\in [\bar d]\times [k],\;\;W_{f(j,m),j}=k^2\cdot e^{\frac{m 2\pi i}{k}}~.$$
We note that for every $\forall j\in [\bar d]$ and $m,m'\in [k]$,
$$(Wx_{j,m})_{f(j,m')}=k^2 \cdot \inner{e^{\frac{m' 2\pi i}{k}},e^{\frac{m 2\pi i}{k}}}
=k^2 \cdot \cos\left(2\pi \frac{f(j,m)-f(j,m')}{k}\right)~.$$
Thus,
$$(Wx_{j,m})_{f(j,m)}\ge k^2\cdot (1-\cos(2\pi/k))+\max_{m'\ne m} (Wx_{j,m})_{f(j,m')}\ge 1+\max_{m'\ne m} (Wx_{j,m})_{m'}$$
Where the last inequality follows from the fact that by Taylor's Theorem, for $x\in [0,2\pi/3]$,
$$1-\cos(x)\ge \frac{x^2}{2}-\frac{x^4}{24}\ge \frac{x^2}{4}$$
\proofbox

\section{Conclusion and future work}\label{sec:conclusion}
We have bounded the price of bandit information in the setting of
hypothesis class based on-line learning and extended the results of \cite{AuerCeFrSc03}. We applied our results to estimate the bandit error rate of the class of large margin classifiers.

The focus of this paper is information theoretic. That is, we have
ignored time complexity issues. It is of interest to study the {\em computational price of bandit information}
-- i.e. how the required runtime grows when moving from the full-info
to the bandit scenario. It is instructive to consider the PAC setting. Given a learning algorithm, $A$, for a class $H$ in the PAC full-info setting we can simply construct a bandit learning algorithm as follows -- given a sample of unlabled instances, we guess, for each instance, a label from $Y$, uniformly at random. Typically, we will be correct on about $\frac{1}{k}$ of the examples. Thus, we can generate a {\em labeled} i.i.d. sample whose size is $\frac{1}{k}$-fraction of the original sample, and run the full-info algorithm $A$ on this sample. Using this construction (see \cite{DanielySaBeSh11}), it easily follows that in the PAC setting, the price of bandit information, both information theoretic and computational, is $O(k)$. Is this true in the on-line setting as well? We note that this question is open and interesting already for the class of large-margin multiclass linear separators.

There is still some room for improvements of the bounds in Theorems \ref{thm:main_realizable} and \ref{thm:main_agnostic}. We conjecture that the optimal bounds are that for every class $H$, $\BErr^r_H(T)=O(k\cdot \Ldim(H))$ and $\BErr^a_H(T)=O(\sqrt{k\cdot \Ldim(H)T})$.

Theorem \ref{thm:main_agnostic} together with Theorem \ref{thm:Littlestone_dimensions} characterize  the bandit-agnostic error rate up to a factor of $\tilde{O}(\sqrt{k})$. It is of interest to find a tighter characterization. We note that Theorem \ref{thm:Littlestone_dimensions} shows that the bandit Littlestone dimension characterizes the error rate in the bandit realizable case for deterministic algorithms. It is an open question to show that this dimension quantifies the error rate also in the agnostic case and for randomized algorithms in the realizable case.

\subsection*{Acknowledgements}
We thank Nati Linial and Shai Shalev-Shwartz for many comments and suggestions regarding this work.
Amit Daniely is a recipient of the Google Europe Fellowship in Learning Theory, and this research is supported in part by this Google Fellowship.
\bibliography{bib}

\appendix
\section{The SOA algorithm}
For completeness, we outline the SOA algorithm (of \cite{Ben-DavidPaSh09} and \cite{DanielySaBeSh11}) for a class $H\subset Y^X$.

\begin{algorithm}[th] 
\caption{Standard Optimal Algorithm (SOA)} \label{algo:SOA}
\begin{algorithmic}[1]
\STATE Initialize: $V_0=H$.
\FOR {$t=1,2, \ldots$}
\STATE receive $x_t$.
\STATE for $y\in Y$, let $V_t^{(y)}=\{f\in V_{t-1}:f(x_t)=y\}$.
\STATE predict $\hat{y}_t\in\argmax_y\Ldim(V_t^{(y)})$.
\STATE receive true answer $y_t$.
\STATE update $V_t=V_t^{(y_t)}$.
\ENDFOR
\end{algorithmic}
\end{algorithm}

\end{document}